\documentclass[sigconf,natbib=True,anonymous=False,authorversion,nonacm]{acmart}

\usepackage{algorithm}
\usepackage{algorithmic}

\usepackage[utf8]{inputenc} 
\usepackage[T1]{fontenc}    
\usepackage{hyperref}       
\usepackage{url}            
\usepackage{booktabs}       
\usepackage{amsfonts}       
\usepackage{nicefrac}       
\usepackage{microtype}      
\usepackage{xcolor}         
\usepackage{enumitem}
\usepackage{dsfont}
\usepackage{hyperref}

\newtheorem*{theorem*}{Theorem}

\setcopyright{none}

\begin{document}

\title{Exceeding the Limits of Visual-Linguistic Multi-Task Learning}

 \author{Cameron R. Wolfe}
 \thanks{Work done while CW was an intern at Salesforce Einstein.}
 \affiliation{%
   \institution{Rice University}
   \streetaddress{}
   \city{Houston,~TX}
   \country{USA}
}
 \email{crw13@rice.edu}

\author{Keld T. Lundgaard}
 \affiliation{%
   \institution{Salesforce Einstein}
   \streetaddress{}
   \city{Cambridge,~MA}
   \country{USA}
}
\renewcommand{\shortauthors}{Wolfe et. al.}

\begin{abstract}
By leveraging large amounts of product data collected across hundreds of live e-commerce websites, we construct 1000 unique classification tasks that share similarly-structured input data, comprised of both text and images.
These classification tasks focus on learning the product hierarchy of different e-commerce websites, causing many of them to be correlated.
Adopting a multi-modal transformer model, we solve these tasks in unison using multi-task learning (MTL).
Extensive experiments are presented over an initial 100-task dataset to reveal best practices for ``large-scale MTL'' (i.e., MTL with $\geq 100$ tasks).
From these experiments, a final, unified methodology is derived, which is composed of both best practices and new proposals such as DyPa, a simple heuristic for automatically allocating task-specific parameters to tasks that could benefit from extra capacity.
Using our large-scale MTL methodology, we successfully train a single model across all 1000 tasks in our dataset while using minimal task specific parameters, thereby showing that it is possible to extend several orders of magnitude beyond current efforts in MTL.
\end{abstract}



\maketitle

\section{Introduction}

\begin{figure}
\hspace{-0.4cm}
\includegraphics[width=3.5in]{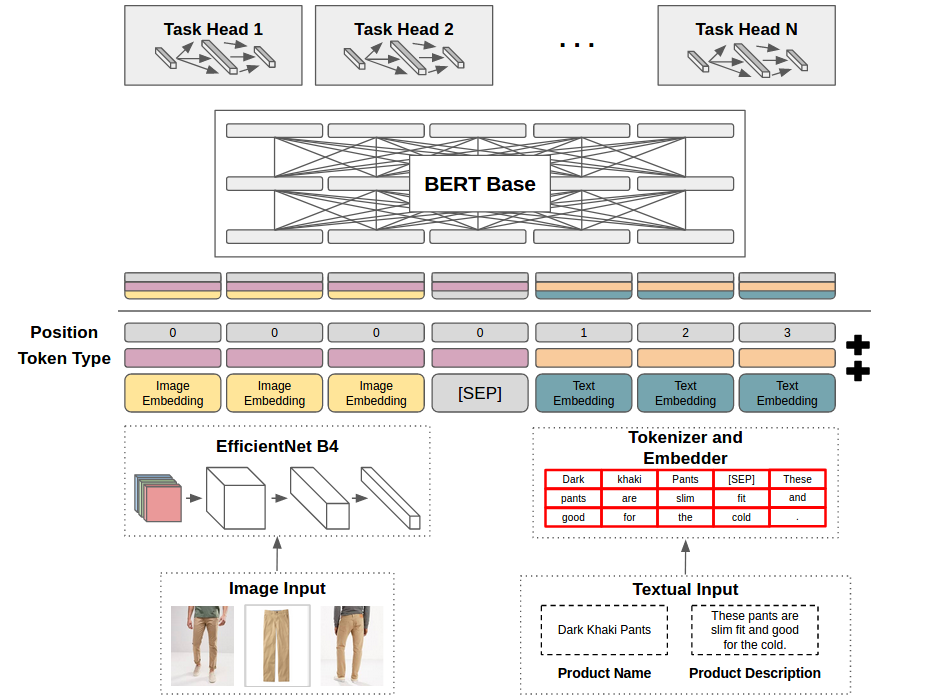}
\caption{A depiction of the BERT-style, multi-modal classification model that was used in all experiments. The main parameters of the network (i.e., the parameters of the transformer) are shared, and a task-specific classification head is assigned to each task.}
\label{fig:model_depict}
\end{figure}

\noindent \textbf{Background.}
The transformer model \cite{transformer} has revolutionized deep learning research \cite{bert, vlp, 16by16words, transobjdet}.
After being proposed for neural machine translation, transformers were adopted for self-supervised training over large language corpora with the proposal of BERT \cite{bert}, allowing models to be pre-trained, made publicly available, and fine-tuned on downstream tasks \cite{bert, roberta}.
Due to their remarkable performance, transformer models became popular in domains beyond natural language processing (NLP), such as computer vision \cite{16by16words, transobjdet}, multi-modal deep learning \cite{vilbert, mmbt}, and even music generation \cite{musictrans}.
The use cases for transformers are constantly expanding as deep learning practitioners develop new applications \cite{biobert, sentence_bert, ernie, video_bert}.

The current trend in transformer research (and deep learning in general) is towards larger models and datasets \cite{gpt3, xlm-r, t5}.
Despite the widespread moral and practical questioning of this trend \cite{costofnlp, sotaaimodels, transcarbonemit, energy_consider}, the deep learning community has continued forward, having yet to reach an upper bound to model capacity and performance \cite{gpt3}.
In fact, significantly overparameterized deep learning models have even been shown to discover high-performing, generalizable solutions \cite{doubledescent}.
These findings illustrate that one may benefit from using a larger model.
However, such benefit comes at the cost of increased inference and training time, possibly to the point of complete intractability \cite{costofnlp, gpt3}.

So, where does this leave deep learning practitioners if they want to reap the performance benefits of larger transformer models?
One option would be to discover low-cost transformer variants that perform well \cite{linformer, funneltransformer, reformer, litetrans, star_transformer, albert}.
Additionally, transformers could be pruned to reduce inference time \cite{layerdrop, powerbert, lth_bert, bert_prune}.
However, despite the valuable contributions of these methodologies, transformer models remain computationally expensive to train, especially if done from scratch.
Inspired by these issues, this work explores an orthogonal direction of research.
Namely, if the computational complexity of the transformer model cannot be completely avoided, \emph{how can we get the most benefit from training a single model?}

Multi-task learning (MTL) with transformer models has been explored by previous work \cite{mtdnn, bert_and_pals, 12in1}.
Because BERT models encode redundant data within their representations \cite{powerbert, bert_loses_patience} and can be pruned without performance deterioration \cite{layerdrop, bert_prune, lth_bert}, such models intuitively contain sufficient capacity to perform well in the MTL domain. 
However, no work has yet fully explored the limits of transformers for MTL (i.e., most works solve only 10-20 tasks in unison \cite{mtdnn, bert_and_pals}).

\noindent \textbf{Our contribution.}
In this work, a multi-modal, multi-task classification dataset is constructed using e-commerce product images, textual descriptions, and categorical attributes.
Such e-commerce data allows for the formation of 1000 similar, but unique classification tasks that can be solved with MTL.
In our case, we focus on groups of task that learn the product hierarchy of different e-commerce websites, which has numerous practical applications in improving personalized product recommendations. 
To handle this dataset, a multi-modal BERT model \cite{mmbt} is adopted and modified to simultaneously solve hundreds, or thousands, of such classification tasks with a single model and minimal task-specific parameters.
We coin this problem ``large-scale MTL'', which represents MTL in the case of $\geq$ 100 tasks being solved by a single model. 
Our work aims to derive a simple, unified approach for training high-performing models within the large-scale MTL domain.
Although we utilize a previously-proposed model architecture, \textit{the novel contribution of this work lies within the training methodology that is developed of large-scale MTL.}

Some notable contributions of this work are as follows:
\begin{itemize}
    \item We perform extensive experiments with an MTL dataset of 100 tasks, resulting in a unified, high-performing methodology for training models in the large-scale MTL domain.
    \item We extensively study the behavior of our proposed methodology, providing numerous ablation experiments for transfer learning, increased model capacity, smaller-scale MTL datasets, and much more.
    \item We demonstrate that our proposed methodology is very scalable and capable of solving 1000 classification tasks simultaneously.
\end{itemize}

\section{Related Work}
Multi-task learning has been a popular topic in deep learning for some time.
Formative works in this area explored methods of optimally weighting tasks within a group or modifying gradients to prevent conflict between tasks \cite{uncertainty_weight, gradient_surgery}.
Following these efforts, numerous methods of handling objectives comprised of multiple tasks were proposed \cite{small_towers, ba_mtl, endtoend_mtl, asymmetric_mtl}.
However, such methods share a common goal of training a unified model over a group of tasks that performs well and limits requirements for task-specific parameters.
Multi-task learning approaches have since been applied to numerous domains, such as forming sentence embeddings \cite{hierarchical_mtl_embedding, ls_mtl_embed}, solving computer vision tasks \cite{uber_net}, and even performing multi-modal reasoning \cite{12in1, omni_net, hierarchical_vl}.
Several, more comprehensive, summaries of developments in the multi-task learning space are also available \cite{ruder_overview, nlp_mtl_overview}.

The introduction of the transformer architecture \cite{transformer} and BERT \cite{bert} revolutionized deep learning for NLP \cite{summarizebert, roberta, bert_transfer, xlm-r} and several other domains \cite{transobjdet, musictrans, biobert, 16by16words}.
Shortly after their introduction, transformer architectures were applied to multi-modal data, leading to numerous variants \cite{pixel_bert, vilbert, vlp, fashion_bert, uniter, image_bert, lxmert, vl_bert, unicoder_vl, visual_bert, b2t2, video_bert}.
Such attention-based approaches for multi-modal learning can be roughly categorized into single-stream and separate-stream methodologies, based on whether all modalities are processed separately or as a single sequence.
Generally, single-stream architectures are popular because of their simplicity and performance \cite{uniter, vlp, mm_pretrn, mm_secrets}.
Many of such single-stream models for multi-modal deep learning share identical architectures to BERT \cite{bert} and can even be initialized with the same pre-trained weights \cite{image_bert, mmbt, mm_pretrn}.
For multi-modal classification, it has been shown that BERT-style, single-stream architectures perform well and are easy to train \cite{mmbt}.

Transformer models, especially variants of BERT, have become popular in the multi-task learning community.
Due to their many parameters (e.g., 110M parameters in BERT base and 340M parameters in BERT large \cite{bert}), these models are capable of learning generalizable representations for many tasks simultaneously.
For example, BERT models have been used to simultaneously solve multiple GLUE tasks \cite{glue, mtdnn, bert_and_pals}, leading to improved generalization across all tasks.
Similarly, multi-task learning was extended to multi-modal applications, leading to the development of visual-linguistic models trained on diverse sets of tasks \cite{hierarchical_vl, 12in1}.
In these cases, multi-task learning was shown to lead to improved overall model performance on most tasks, proving information from other tasks enables positive inductive transfer within a model's shared representations \cite{small_towers}.

\section{Methodology}
\subsection{Dataset}
\label{S:dataset}

\begin{figure}
\hspace{-0.5cm}
    \centering
    \includegraphics[width=3.5in]{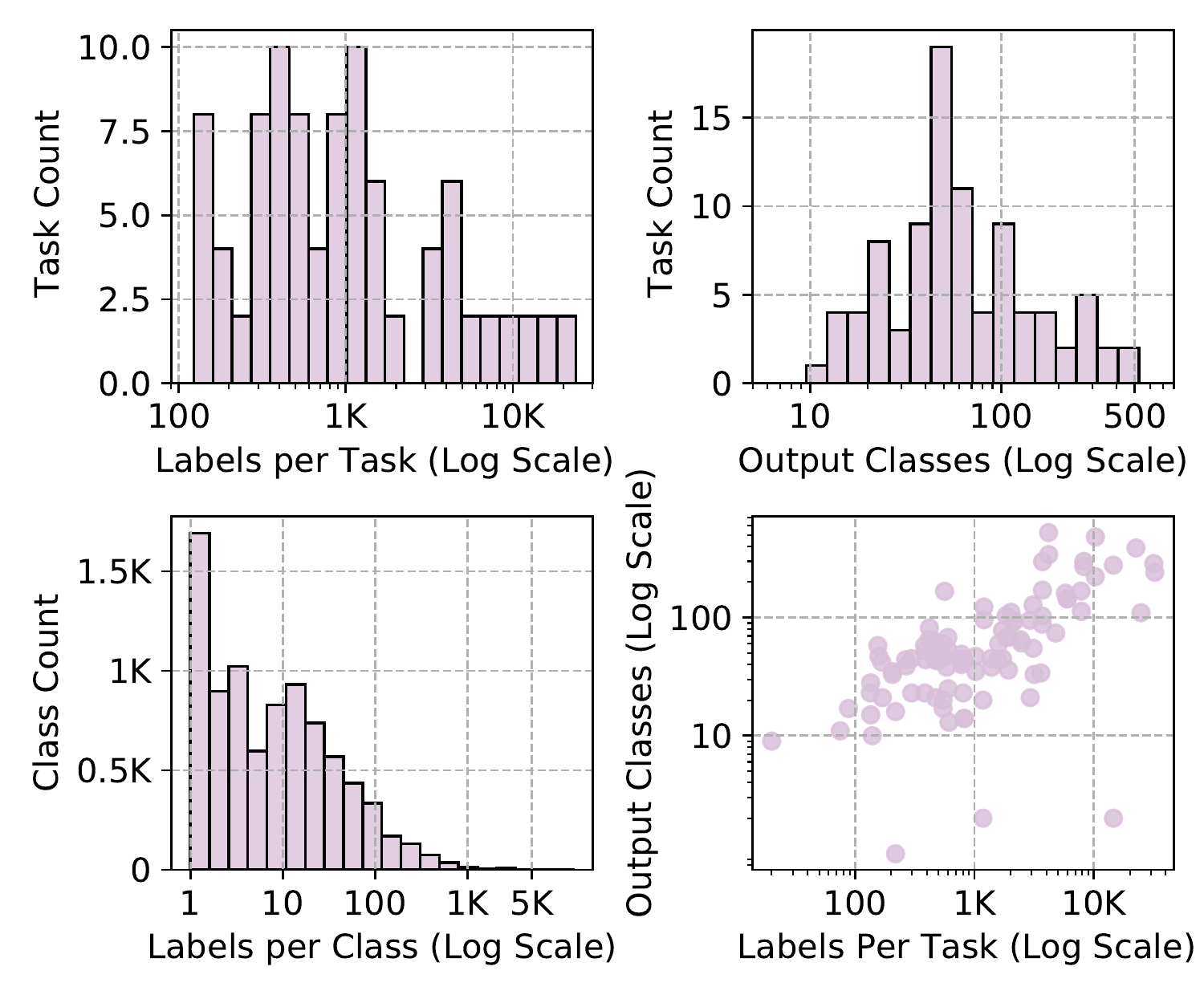}
    \caption{Illustrates various properties of the 100-task dataset including: a histogram of the number of labels for each task (upper left), a histogram of the number of output classes for each task (upper right), a histogram of the number of labels for each output class (lower left), and a scatter plot of the number of labels and output classes for each task (lower right).}
    \label{fig:dataset_metrics}
    \vspace{-0.4cm}
\end{figure}

Our data is obtained from live e-commerce websites and cannot be made publicly available.
However, no multi-modal, multi-task datasets currently exist that provide hundreds of classification tasks with similarly-structured input data (i.e., tasks that can be solved with MTL).
Therefore, this private dataset is the only viable option, to the authors' knowledge, for the study of large-scale MTL.
Nonetheless, we argue that large-scale MTL is quite applicable in the industrial setting.
In our case, such data arises from a multi-tenency of e-commerce websites, but a similar setup could exist for many different software-as-a-service (SaaS) companies with several customers (e.g., web hosting platforms, human resources platforms, advertisement platforms, etc.).
In these cases, developing unified models that generalize across different domains and customers can lower costs and ease the use of such models in practice (i.e., serving a single model for all customers is easier than serving a separate model for every customer). 
In the future, large-scale multi-task datasets could be created or derived from established datasets for further study in this area.

Each example in our dataset, associated with a single e-commerce product, has standardized metadata associated with it, including textual data (e.g., the name or description) and many images. 
Numerous categorical attributes may also be assigned to each product, including the ``type'' and ``category'' attributes.\footnote{Product type corresponds to the lowest-level classification of a product (e.g., shirt, hat, etc.). Product category represents a product's global position in the product hierarchy of a website (e.g., ``Men $\rightarrow$ Dress $\rightarrow$ Shirt'' or ``Children $\rightarrow$ Winter $\rightarrow$ Hat''). The values of these attributes are one-hot encoded to create the set of output classes for a task.}
The possible values for these attributes are not standardized, allowing many unique classification tasks to be created by predicting type and category attributes across different websites.
Though numerous different attributes exist, we choose to study the product type and category attributes because they are related to the product hierarchy of an e-commerce website, which has practical applications in improving product recommendations.
Many of these attribute prediction tasks are difficult to solve in isolation due to a lack of sufficient data (i.e., some websites have very few training examples), making the possibility of positive inductive transfer via MTL appealing.

The dataset used in the majority of experiments, comprised of over 250,000 products, contains 100 attribute prediction tasks, sampled from 50 unique websites (i.e., product type and category prediction for each unique website).
As previously mentioned, both of these attributes are related to the product hierarchy of e-commerce websites.
As a result, many of the attribute prediction tasks in our dataset are correlated (e.g., the product hierarchies of different apparel-based e-commerce websites are usually similar). 
Various metrics and properties of the 100-task dataset are depicted in Fig.~\ref{fig:dataset_metrics}.
As can be seen, there is significant variability in the properties of each task.
Experiments with up to 1000 tasks are also provided in Section~\ref{S:more_task}.
These larger datasets are constructed identically, but with more e-commerce websites. 

\subsection{Model Architecture}
\label{S:model_arch}

We adopt the model architecture proposed in \cite{mmbt} with some slight modifications, outlined below.
The model uses the BERT base uncased architecture and is initialized with the associated pre-trained weights \cite{bert}.
The processing of textual data is identical to that of BERT.
Images are converted to image embeddings with the EfficientNet-B4 model \cite{effnet} and projected to the correct dimension with a learnable linear layer.
Image embeddings are created as a preprocessing step, and the EfficientNet model is not fine-tuned during training.
All image embeddings associated with a product are concatenated with the textual tokens (i.e., divided by a ``[SEP]'' token) to form a single input sequence.
Token type and position embeddings, which are fine-tuned throughout training, are also added to the input sequence.
Image and textual tokens receive different token type embeddings.
The same position embedding is added to all image tokens (i.e., the images are not ordered), but position embeddings are incremented at each position in the textual sequence.
BERT parameters are fine-tuned and shared across tasks.
The output of BERT is passed into a task-specific classification head to perform the final classification, and the correct task must be specified as part of the forward pass.
A detailed illustration of this model is presented in Fig. \ref{fig:model_depict}. 

\subsection{Evaluation} \label{eval}
Because our model is trained to simultaneously solve up to 1000 tasks, evaluating the model's performance is not trivial.
We choose to perform evaluation with several metrics:
\begin{itemize}
    \item \textbf{Mean Accuracy}: The accuracy is computed separately for each task, and the mean of these accuracies is reported.
    \item \textbf{T10 Accuracy}: This metric considers the 10\% of tasks that contain the most number of data examples. The accuracy is computed separately for each of these tasks, and the mean of these accuracies is reported (i.e., mean accuracy of high-resource tasks).
    \item \textbf{B10 Accuracy}: This metric considers the 10\% of tasks that contain the least number of data examples. The accuracy is computed separately for each of these tasks, and the mean of these accuracies is reported (i.e., mean accuracy of low-resource tasks).
\end{itemize}

All of the above metrics are evaluated on hold-out test sets for each task, which are constructed with a uniform 80-20 split.

\subsection{Experimental Details}
Experiments on the 100-task dataset are run for 15 total epochs, where an epoch is defined as a cycle through a number of data examples equal to the size of the full dataset. 
For datasets with more tasks (e.g., see Section~\ref{S:more_task}), the number of epochs is reduced proportionally, based on the number of training examples in the dataset.
All experiments use a batch size of 64, which was the maximum size to fit in a single Tesla V100 GPU.
Although some of the larger tests were run on a100 GPUs with more RAM, we maintain the same batch size across all tests for simplicity.
Each batch is comprised entirely of examples from a single task, and all models are trained with the AdamW optimizer.
Each test was run on a single GPU for simplicity.

\section{How to handle lots of tasks}
\label{S:method_ablations}
In this section, extensive ablation experiments over the 100-task dataset are provided.
The lessons learned from these experiments allowed us to arrive step-by-step at our final methodology, which yields superior performance in the large-scale MTL domain.
\textit{This training methodology, discovered through extensive experimentation and tuning, is the main novel contribution of this work}, as no previous work attempts to solve such a large number of tasks simultaneously using MTL.

\subsection{Optimization}
\begin{figure}
\hspace{-0.4cm}
    \centering
    \includegraphics[width=2.75in]{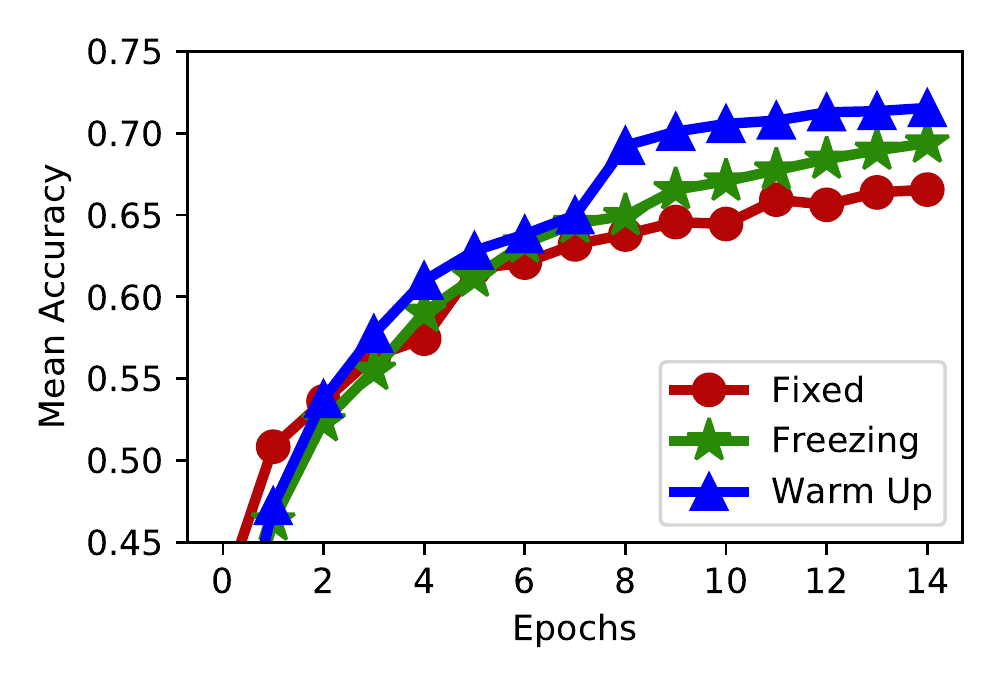}
    \caption{Mean accuracy on the 100-task dataset with different learning rate strategies.}
    \label{fig:opt_acc}
\end{figure}

Because many tasks are being optimized simultaneously, achieving convergence with a large-scale MTL model is not trivial.
In fact, numerous initial attempts at training a model on the 100-task dataset diverged completely.
Eventually, it was discovered that convergence could be achieved by $\emph{i)}$ using a low learning rate throughout training $\emph{ii)}$ freezing the shared transformer weights during initial epochs or $\emph{iii)}$ utilizing a well-tuned learning rate warm up schedule.

The fixed learning rate strategy utilizes a learning rate of $10^{-5}$ throughout training.
The freezing strategy utilizes a higher learning rate of $10^{-4}$ during initial epochs, but keeps the weights of the transformer backbone fixed.
The weights of the transformer are unfrozen after four epochs, at which point the learning rate is reduced by $10\times$.
The warm up schedule is comprised of an initial warm up phase, which increases the learning rate from $10^{-5}$ to $10^{-4}$ during the first four epochs, followed by a step schedule that decreases the learning rate $10\times$ at epochs eight and 12.

A comparison of each of these three optimization strategies is presented in Fig.~\ref{fig:opt_acc}.
Although the fixed learning rate strategy converges, the training process was slow compared to other methods.
The freezing strategy provided a slight improvement in performance, but the best results were achieved using the warm up strategy.
As shown in Fig.~\ref{fig:opt_acc}, the warm up strategy quickly reaches a stable plateau in performance, leading us to adopt this strategy in the following experiments.

\subsection{Task Sampling Strategies}
\begin{figure}
\hspace{-0.4cm}
    \centering
    \includegraphics[width=2.75in]{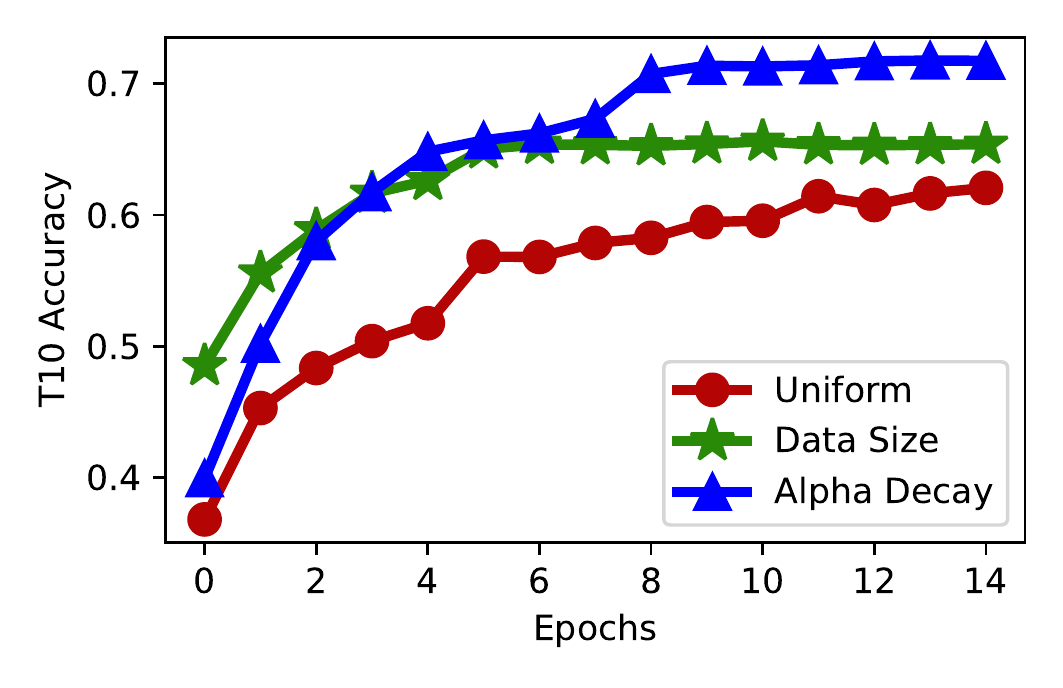}
    \caption{T10 accuracy achieved by models trained with different task sampling strategies on the 100-task dataset.}
    \label{fig:sample_strat}
\end{figure}

Each batch during training contains examples sampled from a single task.
However, the best approach for sampling tasks at each iteration has yet to be standardized, especially in the large-scale MTL domain.
The naive approach would be to randomly sample a task with uniform probability at each iteration.
On the other hand, one could sample tasks with a probability that is proportional to the number of data examples for that task (i.e., referred to as ``data size'' sampling).
These two strategies can also be interpolated to yield the probability of sampling task $T$ for a given batch, $P(T)$, as follows.

\begin{align*}
    P(T) = \frac{N_T^\alpha}{\sum_tN_t^\alpha}
\end{align*}

where $N_T$ is the number of training examples for task $T$ and $\alpha \in [0, 1]$ is a hyperparameter.
The value of $\alpha$ can be simply understood as an interpolation between uniform ($\alpha=0$) and data size ($\alpha=1$) sampling.

Previous work has shown that annealing schedules for $\alpha$ can yield performance improvements for MTL \cite{bert_and_pals}.
To test the validity of such claims for large-scale MTL, we compare decaying the value of $\alpha$ throughout training (i.e., linear decay from 1.0 to 0.0) to data size and uniform task sampling; see Fig.~\ref{fig:sample_strat}.
Although data size sampling seems to provide faster convergence in initial epochs, $\alpha$ decay significantly outperforms both data size and uniform task sampling.
We emphasize that the choice of task sampling strategy significantly impacts performance in the large-scale MTL domain (i.e., 10\% difference in T10 accuracy).
\emph{This observation inspired an extensive exploration of different $\alpha$ decay schedules to ensure the best possible task sampling strategy was discovered.}

\subsection{Choosing an $\alpha$ Decay Schedule}
\begin{table}
\centering
\begin{scriptsize}
\caption{Model performance for different $\alpha$ decay schedules on the 100-task dataset.}\vspace{-0.1cm}
\begin{tabular}{cc|ccc}
\toprule
$\alpha$ Decay Method & $\alpha$ Range & Mean Acc. & T10 Acc. & B10 Acc.\\
\midrule
Linear & 1.0 $\rightarrow$ 0.5 & 74.66\% $\pm$ 0.42 & 73.56\% $\pm$ 0.56 & 50.05\% $\pm$ 1.91 \\
& 1.0 $\rightarrow$ 0.1 & 74.97\% $\pm$ 0.23 & 72.37\% $\pm$ 0.48 & 51.70\% $\pm$ 0.95 \\
& 1.0 $\rightarrow$ 0.0 & 75.31\% $\pm$ 0.31 & 72.49\% $\pm$ 0.74 & 52.98\% $\pm$ 0.29 \\
\midrule
Exponential & 1.0 $\rightarrow$ 0.5 & 74.79\% $\pm$ 0.12 & 73.33\% $\pm$ 0.31  & 51.69\% $\pm$ 0.19 \\
& 1.0 $\rightarrow$ 0.1 & \textbf{75.48\% $\pm$ 0.13} & 71.69\% $\pm$ 0.31  & \textbf{54.09\% $\pm$ 0.88}\\
& 1.0 $\rightarrow$ 0.0 & 74.86\% $\pm$ 0.27 & 69.85\% $\pm$ 1.15 & 52.38\% $\pm$ 0.72 \\ 
\midrule
Cosine & 1.0 $\rightarrow$ 0.5 & 74.63\% $\pm$ 0.11 & 73.71\% $\pm$ 0.27 & 52.25\% $\pm$ 0.27 \\
& 1.0 $\rightarrow$ 0.1 & 74.88\% $\pm$ 0.31 & 73.33\% $\pm$ 0.50 & 53.02\% $\pm$ 0.33 \\
& 1.0 $\rightarrow$ 0.0 & 74.47\% $\pm$ 0.35 & 73.28\% $\pm$ 0.45 & 51.20\% $\pm$ 0.91 \\
\midrule
Demon & 1.0 $\rightarrow$ 0.5 & 73.20\% $\pm$ 1.00 & \textbf{73.82\% $\pm$ 0.69} & 47.46\% $\pm$ 2.90 \\
& 1.0 $\rightarrow$ 0.1 & 74.51\% $\pm$ 0.13 & 73.72\% $\pm$ 0.41 & 52.01\% $\pm$ 1.51 \\
& 1.0 $\rightarrow$ 0.0 & 74.31\% $\pm$ 0.10 & 73.84\% $\pm$ 0.26 & 52.92\% $\pm$ 0.81 \\
\bottomrule
\end{tabular}\vspace{-0.1cm}
\label{alpha_decay_table}
\end{scriptsize}
\end{table}

\begin{figure}
\hspace{-0.2cm}
    \centering
    \includegraphics[width=2.75in]{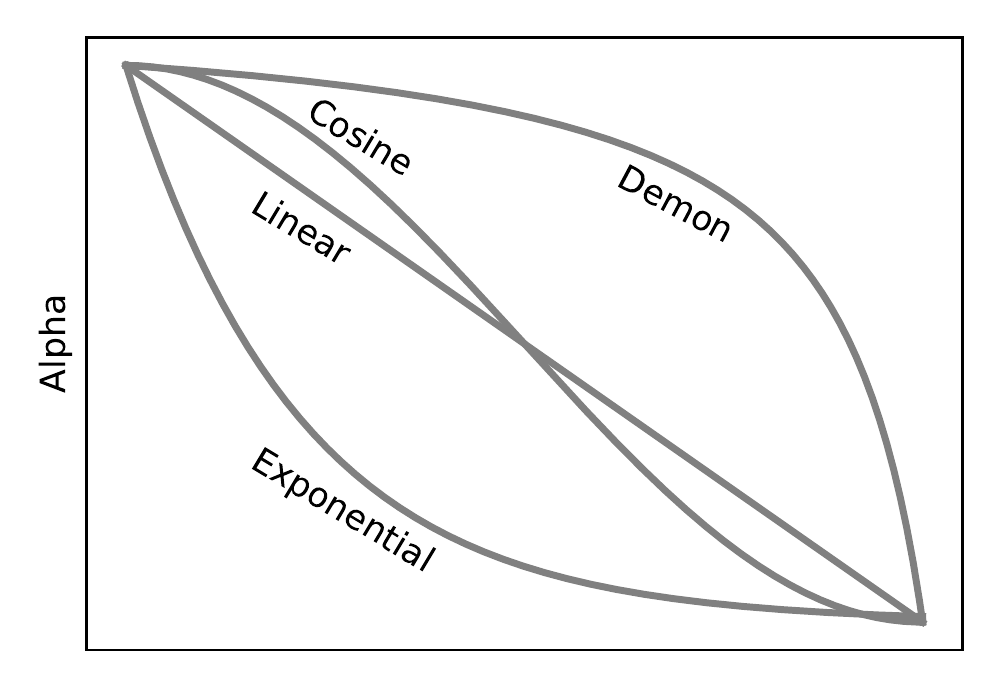}
    \caption{Illustration for all different $\alpha$ decay schedules that were tested. All decay schedules started and ended at the beginning and ending of training, respectively. The \texttt{Demon} schedule was adopted from \cite{demon}.}
    \label{fig:decay_schedules}
\end{figure}

We empirically compare numerous options for $\alpha$ decay, depicted in Fig. \ref{fig:decay_schedules}.
Exponentially decaying $\alpha$ from a value of 1.0 to 0.1 throughout training was found to consistently achieve balanced performance; see Table \ref{alpha_decay_table}.
As a result, we adopt this approach in the rest of experiments.
Intuitively, the exponential $\alpha$ decay schedule works well because it performs data size sampling during the early parts of training, where it is conducive to faster convergence.
However, the schedule quickly decays to lower $\alpha$ values to avoid damaging the model's performance.
Experiments were also performed using restarts, warm up, and cyclical schedules for $\alpha$, but none of these more complex schedules improved performance.

\subsection{How often should a new task be sampled?}

Previous work has shown that performing consecutive iterations on the same task could degrade MTL performance \cite{bert_and_pals, mtdnn}.
We find that the same is true within the large-scale MTL domain.  
Interestingly, performing as few as ten consecutive iterations on each task during training led our model to completely diverge on the 100-task dataset.
Therefore, \emph{sampling a new task at every iteration of training is seemingly conducive to good performance.}

\subsection{Limiting Task Specific Parameters}
\label{S:param_eff}
\begin{figure}
\hspace{-0.4cm}
    \centering
    \includegraphics[width=2.75in]{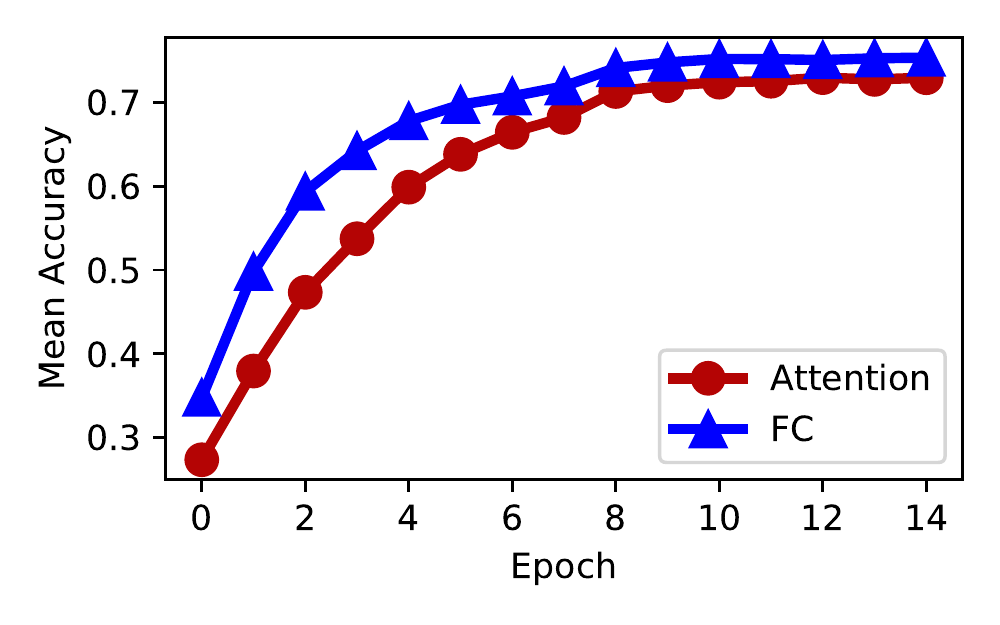}
    \caption{Comparison of the performance of attention-based task-specific heads to FC task-specific heads in terms of mean accuracy.}
    \label{fig:pal_vs_linear}
\end{figure}

Within our implementation, each task must have a task-specific classification head.
A ``naive'' classification head implementation -- a fully-connected (FC) layer followed by a nonlinearity (e.g., ReLU) -- consumes many parameters.
For example, FC classification heads consume 58 million parameters for the 100-task dataset.\footnote{Fewer parameters could be potentially consumed by performing a single linear projection from the tokens outputted by BERT to the output space. However, we found that such an approach performed worse than utilizing a classification head that contains a non-linearity.}
Intuitively, this problem worsens as the methodology scales to hundreds, or thousands, of tasks.
Therefore, \emph{a more parameter-efficient approach had to be developed to make large-scale MTL more feasible.}

Inspired by \cite{bert_and_pals}, we formulate each task-specific classification head as a low-dimension self-attention module. 
More specifically, we take the final token sequence outputted from BERT, project the tokens to some lower dimension $d_t$, then perform self-attention at this lower dimension.
After the self-attention layer, tokens are averaged and passed through a linear classification layer to get the final output.
Using this formulation, we are able to significantly decrease usage of task-specific parameters (e.g., $d_t = 64$ uses $10\times$ fewer parameters than FC classification heads).
The performance of FC and attention-based classification heads (i.e., with $d_t = 64$) is shown in Fig. \ref{fig:pal_vs_linear}.
As can be seen, the attention-based task-specific heads, despite using $10\times$ fewer parameters, maintain similar performance to the FC classification heads, leading us to use them within the proposed methodology.

\subsubsection{Investing Task-Specific Parameters Wisely}
\label{S:invest_param}

Though parameter-efficiency makes large-scale MTL more feasible, increasing the number of task-specific parameters may be useful if performance is improved.
To investigate whether performance can be improved with more task-specific parameters, we run experiments using task-specific attention modules with different numbers of heads and hidden dimensions.\footnote{Task-specific attention modules with multiple layers were also tested, but led to no noticeable performance improvement.}
The results of these experiments are presented in Table~\ref{attention_sizes_table}.
Assuming a fixed value of $d_t$, changing the number of attention heads does not significantly impact performance.
Furthermore, model performance on high-resource tasks consistently improves as $d_t$ is increased, but B10 accuracy begins to degrade as $d_t$ becomes too large, revealing that larger classification heads cause overfitting on low-resource tasks.
Because the sizes of task-specific heads cannot be tuned by hand in the large-scale MTL domain (i.e., this would require manually setting $d_t$ for hundreds or thousands of tasks), \emph{there is a need for some heuristic that automatically allocates task-specific parameters to tasks that need them}.

\begin{table}
\centering
\begin{scriptsize}
\caption{Performance of different sizes of task-specific classification heads on the 100-task dataset. For DyPA, the hidden dimension of task-specific attention heads in the lowest quartile is listed. This dimension increases by $2\times$ in each following quartile.}\vspace{-0.1cm}
\begin{tabular}{ccc|ccc}
\toprule
Task-Specific Head Type & \# Attn. Heads & $d_t$ & Mean Acc. & T10 Acc. & B10 Acc. \\
\midrule
Fixed-Size & 2 & 64 & 72.95\% & 66.39\% & \textbf{55.07\%} \\
& 4 & 64 & 72.28\% & 66.41\% & 51.29\%\\
& 2 & 128 & 72.17\%  & 69.55\% & 54.41\%  \\
& 4 & 128 & 72.28\% & 68.85\% & 52.04\%  \\
& 2 & 512 & 72.04\% & 67.11\% & 51.79\%  \\
& 4 & 512 & 73.74\% & 68.70\% & 51.43\%  \\
& 2 & 1024 & 73.69\% & 69.72\%  & 51.06\%  \\
& 4 & 1024 & 73.69\% & 69.72\%  & 51.05\%  \\
\midrule
DyPA & 2 & 128 ($2\times$) & \textbf{74.33\%} & \textbf{69.83\%} & 54.39\% \\
& 4 & 128 ($2\times$) & 74.18\% & 69.77\% & 54.00\% \\
\bottomrule
\end{tabular}\vspace{-0.1cm}
\label{attention_sizes_table}
\end{scriptsize}
\end{table}

\subsection{Dynamic Parameter Allocation (DyPA)}
Dynamic Parameter Allocation (DyPA) is a heuristic we propose to solve the problem outlined in Section~\ref{S:invest_param}.
The intuitive idea behind DyPA is to dynamically provide more task-specific parameters to tasks that would benefit from increased capacity (i.e., complex or difficult tasks).
In DyPA, the number of labels associated with each task is used as a proxy for its complexity.
The number of labels for each task is first normalized with respect to that of all other tasks, providing a normally-distributed complexity score for each task.
Using this score, all tasks are separated into quartiles.
Then, the size of a task's classification head is determined by the quartile in which it resides.
Tasks in higher quartiles are given larger task-specific classification heads (i.e., a larger value of $d_t$) and vice versa.
We implement DyPA using attention-based classification heads (see Section \ref{S:param_eff}), but the scheme can be trivially extended to other task-specific module variants.

DyPA introduces a few new hyperparameters, as one must decide the sizes of task-specific heads to be used in the first and last quartiles.\footnote{$d_t$ must be specified for the first and fourth quartiles, then the values of $d_t$ for the second and third quartiles can be linearly interpolated.} 
We observe that T10 accuracy begins to saturate when $d_t > 1024$.
Therefore, we set $d_t = 128$ and $d_t = 1024$ for the first and fourth quartiles, respectively (i.e., quartiles two and three then have sizes 256 and 512).
Using these settings, DyPA is tested with both two and four attention heads.
As can be seen in Table \ref{attention_sizes_table}, DyPA improves mean and T10 accuracy in comparison to fixed-size classification heads, while maintaining comparable B10 accuracy.
In other words, \emph{DyPA improves performance on high-resource tasks without overfitting on low-resource tasks.}
Furthermore, DyPA still uses roughly $3.5\times$ fewer task-specific parameters in comparison to FC classification heads.

\section{Results}
\label{S:result}
\begin{table}
\centering
\begin{scriptsize}
\caption{Performance of the proposed methodology on datasets containing different numbers of tasks. Baseline performance is obtained by training a separate model on each task in the 100-task dataset. All metrics are evaluated over the 100-task dataset.}\vspace{-0.1cm}
\begin{tabular}{c|ccc}
\toprule
Method & Mean Acc. & T10 Acc. & B10 Acc.\\
\midrule
Baseline & 61.50\% & \textbf{72.33\%} & 34.40\% \\
100 Task MTL & 73.29\% $\pm$ 0.43 & 69.30\% $\pm$ 0.53 & 54.39\% $\pm$ 0.42 \\
500 Task MTL & 72.54\% $\pm$ 0.13 & 67.36\% $\pm$ 0.13 & 52.80\% $\pm$ 1.80 \\
1000 Task MTL & \textbf{74.98\% $\pm$ 0.11} & 67.02\% $\pm$ 0.15 & \textbf{57.08\% $\pm$ 0.65}\\
\bottomrule
\end{tabular}\vspace{-0.1cm}
\label{final_results_table}
\end{scriptsize}
\end{table}

In this section, the performance of our final large-scale MTL methodology, outlined in Section~\ref{S:method_ablations}, is evaluated over the 100-task dataset and larger datasets.
As a baseline, we train models separately on each task in the 100-task dataset.
It should be noted that this baseline is comprised of 100 BERT base models and is nearly infeasible to use in practice.
Because our training methodology for large-scale MTL incorporates best practice of numerous previously-proposed training methodologies (see Sec. \ref{S:method_ablations}), we do not directly adopt such methods as baselines within this section.
Rather, we refer to the extensive ablation experiments in Sec. \ref{S:method_ablations} for comparisons to alternative MTL training methodologies and choose not to replicate such comparisons on larger datasets due to extensive compute requirements.
We demonstrate that our methodology effectively trains models to achieve high performance on all tasks, \emph{even when the number of tasks is increased to as many as 1000.}

\subsection{100-Task Dataset}

We first evaluate the proposed methodology on the 100-task dataset.
As can be seen in Table \ref{final_results_table}, MTL improves mean and B10 accuracy by over 10\% and 20\% in comparison to the baseline, respectively.
Although individually-trained models achieve slightly improved T10 accuracy, such performance is to be expected, as high-resource tasks contain sufficient training data to perform well in isolation (i.e., each task in the T10 accuracy group has $>10$K training labels).
In the low-resource scenario, however, the baseline performance is quite poor (i.e., $<35\%$ B10 accuracy).
The positive inductive transfer provided by large-scale MTL improves performance on these low-resource tasks dramatically (i.e., over 20\% absolute improvement), while only decreasing T10 accuracy by 3\%.
Furthemore, the average performance of MTL models across all tasks in the 100-task dataset is markedly increased in comparison to the baseline, improving by nearly 12\%.
Therefore, these results indicate that our proposed methodology,  \emph{in addition to allowing 100 BERT base models to be compressed into a single model}, provides a significant performance benefit on nearly all tasks.

\subsection{More Tasks}
\label{S:more_task}
To test the ability of the proposed methodology to generalize to larger numbers of tasks, we train our model on datasets with up to 1000 tasks.
As explained in Sec. \ref{S:dataset}, these larger datasets are constructed identically to the 100-task dataset, but with more e-commerce sites.
To make the performance of these models comparable to the baseline, we perform evaluation only over the original 100 tasks, and use the remaining tasks only for training purposes.

As seen in Table \ref{final_results_table}, our proposed large-scale MTL methodology performs well as the number of tasked is scaled well beyond 100.
It should be noted that such performance is only measured over the original 100 tasks, and these models are still solving hundreds of tasks on top of the original 100. 
Although the 500-task model yields slightly degraded performance, all of its validation metrics are within 2\% of those of the 100-task model, revealing that its results are comparable.
Furthermore, the 1000-task model, despite yielding slightly degraded T10 accuracy, outperforms the 100-task model on other metrics.
In particular, the 1000-task model achieves the best mean and B10 accuracy, yielding 1.7\% and 2.7\% absolute improvements over the 100-task model, respectively.
Such a result indicates that, even as the number of tasks is increased to as many as 1000, our proposed methodology can continue to benefit from positive inductive transfer between tasks.
As expected, this positive inductive transfer seems to be most beneficial for low-resource tasks, as the 1000-task MTL model is shown to outperform all other models in B10 accuracy by a significant margin.

Models trained on the larger MTL datasets continue to significantly outperform the baseline in terms of mean and B10 accuracy. 
Most notably, the 1000-task model improves upon the B10 accuracy of the baseline by 22.7\%, highlighting that our methodology's massive benefit to low-resource tasks improves as the number of tasks is increased. 
In addition, the 500 and 1000-task models both outperform the baseline by over 10\% in terms of mean accuracy.
As with the 100-task model, the baseline achieves better T10 accuracy in comparison to MTL models trained on larger datasets.
However, it should be noted that, for 500 and 1000-task models, the baseline could not even be constructed, as it would require the training of several hundred BERT base models.
In the 1000-task case, \emph{such a baseline would consume over 110 billion parameters, while our model consumes roughly 250 million parameters.}

\section{Ablation Experiments}
\subsection{Transfer Learning}
\label{S:tlearn}

\begin{table}
\centering
\begin{scriptsize}
\caption{Classification performance for models initialized with either pre-trained BERT weights or large-scale MTL weights, then fine-tuned on a downstream classification task.}\vspace{-0.1cm}
\begin{tabular}{c|c}
\toprule
Initialization Method & Test Accuracy \\
\midrule
BERT Base & 90.27\% $\pm$ 0.002\\
MTL (100 Task) & 90.77\% $\pm$ 0.001\\
MTL (1000 Task) & \textbf{91.12\% $\pm$ 0.075}\\
\bottomrule
\end{tabular}\vspace{-0.1cm}
\label{tlearn_table}
\end{scriptsize}
\end{table}

Similarly to pre-trained transformer models, our large-scale MTL models are trained over large datasets to simultaneously perform several tasks.
Intuitively, then, the representations learned through large-scale MTL should be useful for transfer learning purposes.
To test this theory, we fine-tune models that are ``pre-trained'' with large-scale MTL on a separate e-commerce classification dataset. 
This dataset contains 405,840 examples and 2,196 output classes.
The data is formatted as described in Section \ref{S:dataset}, but consists of examples across numerous e-commerce websites.
In other words, it is a unified ontology of product categories across many e-commerce websites.

The model used for fine-tuning is the same as shown in Fig. \ref{fig:model_depict}, but it has only a single classification head.
The transformer backbone of this model is initialized using either BERT base weights or the parameters of a large-scale MTL model (i.e., both the 100 and 1000-task models are tested).
The entire model is then fine-tuned for 10 epochs on the downstream ontology classification task.\footnote{Learning rate and weight decay are selected with a grid search using a hold-out validation set. Then, the model is retrained on the full dataset with optimal hyperparameters to measure performance on the test set.}
The performance of the fine-tuned models on a hold-out test set, averaged across three separate trials, is shown in Table \ref{tlearn_table}.
As can be seen, initializing the model with weights learned through large-scale MTL yields improved downstream performance.
This performance improvement becomes more noticeable as the number of tasks used during large-scale MTL training is increased.
In particular, initializing with the 100-task and 1000-task MTL models yields a 0.50\% and 0.85\% increase in performance relative to initializing the model with pre-trained BERT base parameters. 
Given the extensive pre-training of BERT base, this consistent, noticeable improvement in transfer learning performance is quite surprising (i.e., our 100-task model is trained on a single V100 GPU in less than one day) and demonstrates the value of representations learned through large-scale MTL.

\subsection{Increasing Model Capacity}
\begin{table}
\centering
\begin{scriptsize}
\caption{Comparison of MTL models trained on the 100-task dataset with both BERT base and BERT large used as transformer backbones.}\vspace{-0.1cm}
\begin{tabular}{c|cccc}
\toprule
Model Backbone & Mean Acc. & T10 Acc. & B10 Acc.\\
\midrule
BERT Base & 73.29\% $\pm$ 0.43 & \textbf{69.30\% $\pm$ 0.53} & \textbf{54.39\% $\pm$ 0.42} \\
BERT Large & \textbf{73.39\% $\pm$ 0.39} & 68.81\% $\pm$ 0.64 & 52.24\% $\pm$ 0.93 \\
\bottomrule
\end{tabular}\vspace{-0.1cm}
\label{tab:bert_large}
\end{scriptsize}
\end{table}

The proposed model architecture, as shown in Fig. \ref{fig:model_depict}, uses BERT base as a shared backbone between all tasks.
Intuitively, increasing the capacity of this transformer backbone could yield improved model performance.
To test this, the BERT base backbone is replaced with BERT large \cite{bert}, and we train this larger model on the 100-task dataset.
The results obtained with the BERT large backbone are shown in Table \ref{tab:bert_large}.
The larger model obtains slightly improved mean accuracy, but is outperformed by BERT base on other metrics.
Because increasing the capacity of the transformer backbone does not yield noticeably improved performance, these experiments clearly demonstrate the shocking capacity of the BERT base model in the large-scale MTL domain.
Namely, BERT base has seemingly sufficient capacity to perform well across hundreds of tasks simultaneously, even in comparison to a model containing $4\times$ more parameters.

\subsection{Robustness to Larger Tasks}
\begin{figure}
\hspace{-0.4cm}
    \centering
    \includegraphics[width=2.75in]{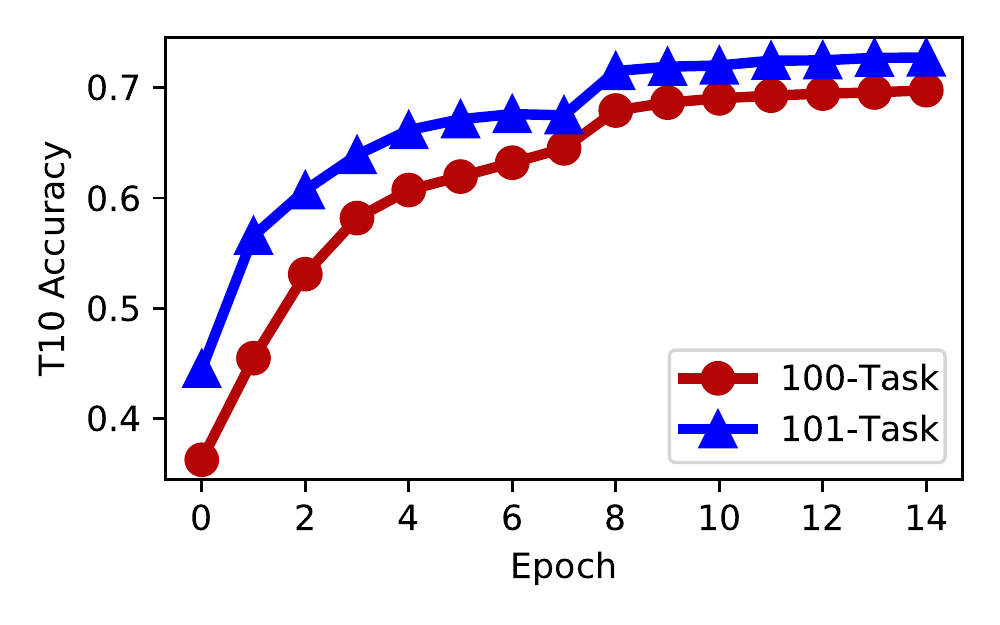}
    \caption{Performance of the proposed methodology when an extra, extremely large task is added to the 100-task dataset.}
    \label{fig:large_task}
\end{figure}

Several aspects of our proposed methodology depend upon the distribution over the number of labels for each task (e.g., $\alpha$-decay and DyPA).
Consequently, one could question whether the model's performance would be sensitive to the properties of this distribution.
For example, adding an extra, very large task into the set of possible tasks could scew the distributions used for sampling and binning within $\alpha$-decay and DyPA, leading to a noticeable performance degredation.
To test this, we add the ontology classification dataset described in Sec.~\ref{S:tlearn} as another task within our 100-task dataset, which is an order of magnitude larger than any other task (see Fig.~\ref{fig:dataset_metrics}).
The results obtained by training a large-scale MTL model on this 101-task dataset are shown in Fig. \ref{fig:large_task}.
As can be seen, the proposed methodology is robust to the addition of this larger task.
In fact, the addition of the ontology classification task results in a 4\% absolute improvement in T10 accuracy.
Therefore, it is clear that our large-scale MTL methodology benefits from the addition of more and larger tasks.

\subsection{From Small to Large Scale MTL}

\begin{table}[]
    \centering
    \begin{tabular}{c|cc|c}
    \toprule
        \# Tasks & Use $\alpha$-decay & Use DyPA & Mean Accuracy\\
        \midrule
        10 & $\times$ & $\times$ & 60.90\%\\
        & $\checkmark$ & $\times$ & 67.46\% \\
        & $\checkmark$ & $\checkmark$ & \textbf{69.97\%}\\ \midrule
        25 & $\times$ & $\times$ & 73.42\%\\
        & $\checkmark$ & $\times$ & 76.25\% \\
        & $\checkmark$ & $\checkmark$ & \textbf{78.12\%}\\ \midrule
        50 & $\times$ & $\times$ & 67.71\%\\
        & $\checkmark$ & $\times$ & 75.12\% \\
        & $\checkmark$ & $\checkmark$ & \textbf{75.47\%}\\ \midrule
        75 & $\times$ & $\times$ & 71.40\%\\
        & $\checkmark$ & $\times$ & 77.34\% \\
        & $\checkmark$ & $\checkmark$ & \textbf{78.65\%}\\
        \bottomrule
    \end{tabular}
    \caption{Experiments that display the effectiveness of our proposed MTL methodology at numerous different scales. DyPA and $\alpha$-decay are selectively eliminated from training to observe their respective performance benefits.}
    \label{tab:small_scale}
\end{table}

Although we have shown that the proposed methodology is useful in the large-scale MTL domain, one may ask whether a similar approach is beneficial with fewer tasks.
To study our methodology's impact at smaller scales, we measure performance of MTL models trained across 10, 25, 50, and 75 tasks (i.e., smaller sub-samplings of the 100-task dataset).
Furthermore, both $\alpha$-decay and DyPA are selectively eliminated from training to directly observe their benefit on the model's performance.
For each of these experiments, only mean accuracy is reported for simplicity.

The results of these smaller-scale MTL experiments are shown in Table \ref{tab:small_scale}.
As can be seen, even when fewer tasks are used during training, employing both $\alpha$-decay and DyPA always leads to the best performance.
Interestingly, the gap in performance between the vanilla model (i.e., trained without $\alpha$-decay or DyPA) and the proposed methodology is relatively similar at all scales (i.e., 4.5-9.0\% improvement across datasets).
As a result, our methodology is shown to be generally useful for MTL, agnostic of the number of tasks available for training.

\section{Conclusions and Future Work}
In this work, we provide proof that overparameterized transformer models are capable of achieving more than previously thought possible.
Using a large-scale MTL dataset comprised of 100 tasks, extensive experiments are performed that allow us to understand how multi-modal BERT models can be most easily trained to simulataneously solve hundreds of tasks.
In particular, we propose a novel DyPA scheme to limit the number of task-specific parameters used by our model and discover best-practices for task sampling within large-scale MTL, \textit{resulting in a final training methodology for large-scale MTL that is the main contribution of this work}.
We then demonstrate that our large-scale MTL training approach is scalable, capable of being trained across 1000 tasks.
By showing what is possible with a single model, we hope this work inspires others, both in research and industry, to explore new, creative methods of using large transformer models efficiently.
In the future, we hope to extend our methodology towards learning different types of tasks and doing zero-shot learning on new tasks.
In addition, we think that it would be very worthwhile to explore 
the robustness of large-scale MTL to adversarial tasks, which
could increasingly be a problem as you extend the number and
variety of tasks.

\bibliographystyle{ACM-Reference-Format}
\bibliography{sample-base}


\end{document}